\documentclass{article}

\usepackage{microtype}
\usepackage{graphicx}
\usepackage{subfigure}
\usepackage{soul}
\usepackage{stfloats}
\usepackage[dvipsnames]{xcolor}
\usepackage{booktabs} 
\usepackage{comment}

\usepackage{hyperref}

\usepackage[preprint]{icml2021}

\icmltitlerunning{ZTop: Zero Training Overhead Portfolios}

\begin{document}

\twocolumn[

           
 \icmltitle{
 Zero Training Overhead Portfolios \\
 for
 Learning to Solve Combinatorial Problems 
      }     
        
         




\begin{icmlauthorlist}
\icmlauthor{Yiwei Bai}{cornell}
\icmlauthor{Wenting Zhao}{cornell}
\icmlauthor{Carla P. Gomes}{cornell}

\end{icmlauthorlist}

\icmlaffiliation{cornell}{Department of Computer Science, Cornell University, Ithaca, USA}

\icmlcorrespondingauthor{Yiwei Bai}{yb263@cornell.edu}

\icmlkeywords{Machine Learning, ICML}

\vskip 0.3in
]



\printAffiliationsAndNotice{}  

\begin{abstract}
There has been an increasing interest in harnessing deep learning to tackle combinatorial optimization (CO) problems in recent years. 
Typical CO deep learning approaches leverage the problem structure in the model architecture. 
Nevertheless, the model selection is still mainly based on the conventional machine learning setting.
Due to the 
discrete nature of CO problems, 
a single model is unlikely to learn the problem entirely. We introduce \textbf{ZTop}, which stands for Zero Training Overhead Portfolio, a simple yet effective model selection and ensemble mechanism 
for learning to solve combinatorial problems. 
\textbf{ZTop}  is inspired by algorithm portfolios, a popular CO ensembling strategy, particularly  restart portfolios,
  which periodically restart a randomized CO algorithm, de facto exploring the search space with different heuristics.
We have observed that well-trained models acquired in the same training trajectory,
with similar top validation performance, perform well on very different validation instances.
Following this observation, 
\textbf{{ZTop}} ensembles a set of well-trained models, each providing a unique heuristic with \emph{zero training overhead},
and 
applies them, sequentially or in parallel, to solve the test instances. 
We show how  \textbf{ZTopping,} i.e., using a ZTop ensemble strategy with a given deep learning approach, can significantly improve the performance of the current state-of-the-art deep learning approaches 
on three  
prototypical CO domains, the hardest unique-solution Sudoku instances, challenging routing problems, and the graph maximum cut problem, as well as on multi-label classification, a 
machine learning task with a large combinatorial label space.
\end{abstract}

\section{Introduction}
Deep learning has achieved tremendous success in many areas, including visual object recognition, neural machine translation, and autonomous driving. 
In contrast, combinatorial optimization problems are challenging  for deep learning given that, in general, they are unsupervised or weakly supervised and have a large combinatorial solution space \cite{coreview, attention}.
A classical approach to solving CO problems is to design useful heuristics manually to guide the algorithm on exploring the large combinatorial search space.
 Nevertheless, a heuristic is typically specialized for a specific problem, and it is not trivial to adapt to other CO problems. 
Deep reinforcement learning (DRL) is becoming a go-to approach for learning heuristics for CO problems  \cite{pointerrl, rl4coreview}. The hope is that DRL  generalizes well to tackle several CO problems  by learning heuristics from scratch. 
The architecture design of a deep learning framework leverages the specific problem structure, e.g., it employs the attention mechanism for routing problems \cite{attention}. However, the model selection method still follows the conventional machine learning style, i.e., it selects
only the model with the best validation performance from the same training trajectory.
Due to the problems' combinatorial nature, we conjecture that a single model typically is not enough to capture the entire problem structure well and learn a useful heuristic for solving the potentially very different problem instances. We propose a new model selection and ensemble method to tackle this challenge.
\begin{figure}[t!]
\centering
\includegraphics[width=0.35\textwidth]{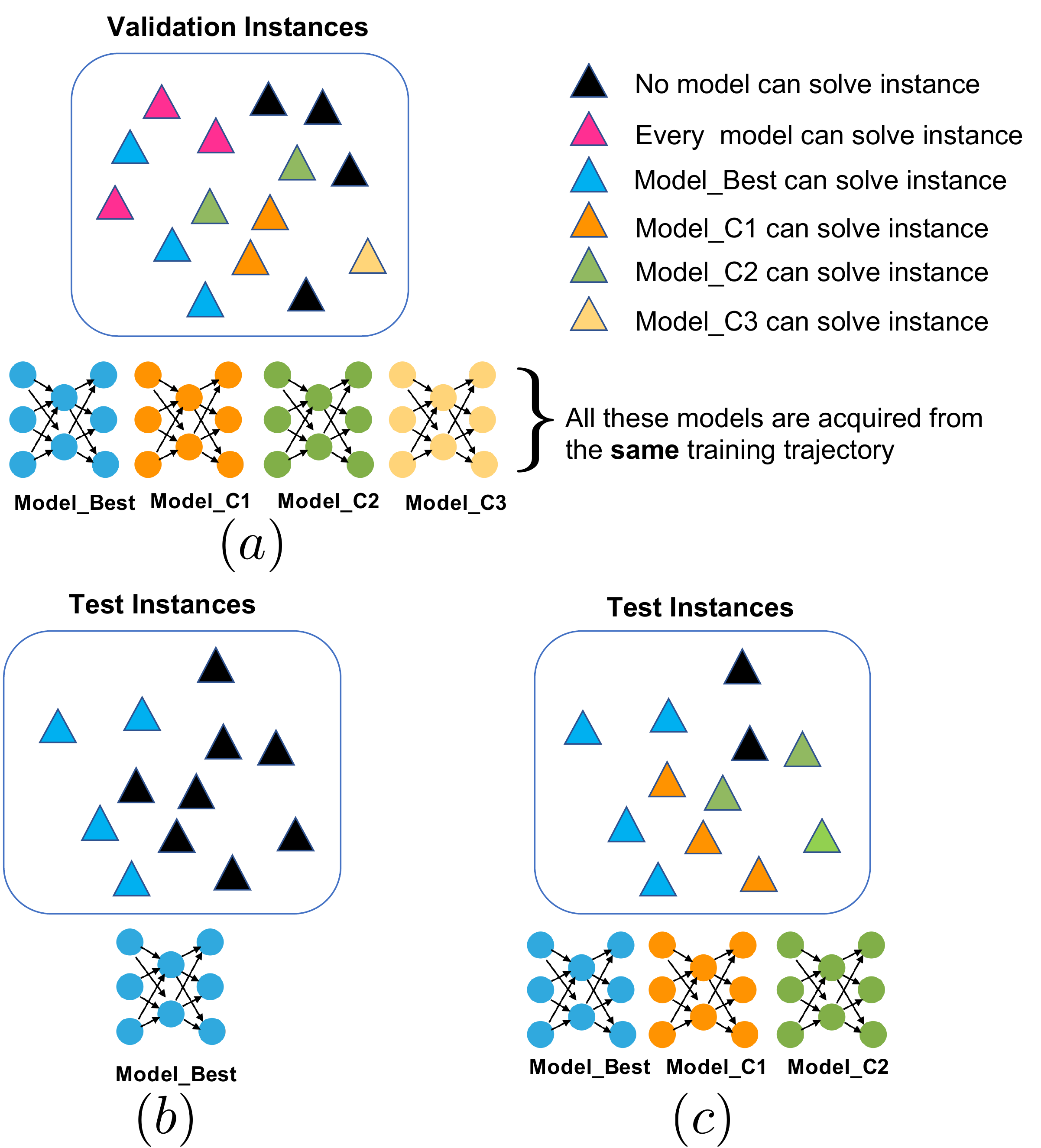}
\caption{\textbf{High-level concept of the \textbf{ZTop} ensemble method.} Assume we use a  deep learning model to solve Sudoku. The top rectangle represents  Sudoku validation instances. A model (heuristic) can solve the 
validation 
instances with the same color. All the well-trained models can solve pink instances. No model can solve black instances. Specific models (other colors) can solve other instances  (a) The model selection mechanism of \textbf{ZTop}. We observe that well-trained models (i.e., with  similar validation performance) acquired in the same training trajectory can solve significantly diverse instances.  (b) Traditional model selection method, i.e., selecting the model with the best validation performance. Due to the combinatorial nature  of CO problems, a single model typically cannot learn the problem entirely, so it can only solve few instances. (c) The \textbf{ZTop} method selects several additional models with zero training overhead. These models have similar validation performance but the test instances they can solve are quite different. Thus, applying these models sequentially or in parallel to solve instances and taking the best output can work well for most instances.}
\label{fig:illustration}
\end{figure}

The runtime and performance of combinatorial optimization algorithms for a given problem can vary significantly from instances to instances, depending on the heuristics used. Even when solving the same instance, a randomized heuristic can also vary dramatically. In fact, the run time distributions of combinatorial optimization algorithms often exhibit heavy tails~\cite{phase}.  To remedy and even exploit the heavy-tailed phenomena and large performance variance of CO search methods, \textbf{\textit{algorithm portfolios}}, and in particular  \textbf{\textit{restart portfolios}}, are widely used in the CO community~\cite{gomes2001algorithm,satzilla,restart, Biere2018}. Essentially, in  restart portfolios,  the CO algorithms are periodically restarted to explore different search space parts, using randomized or pseudo-randomized heuristics and caching useful run information. De facto \textbf{\textit{restarts}} correspond to using thousands of  heuristic variants, with low heuristic switching costs.

We have also observed a large variance in the deep learning models' performance when considering CO problems.  In a deep learning setting, we acquire lots of  models during the same training trajectory, and each model can be viewed as a heuristic. However, using thousands of models in the test phase would be computationally intractable. Our goal is to select a few models covering very diverse heuristics.

\textbf{Our Contributions}:  \textbf{(1)} \textbf{ZTop}, a simple yet \textbf{\textit{effective mechanism to select a diverse set of  models covering different heuristics}} with \textbf{\textit{zero training overhead}} (see Fig.~\ref{fig:illustration}).
The basic idea is to find a few models from the same training trajectory with similar \textbf{\textit{average}} performance (near the best model's performance) but with 
a high variance with respect to the instances they can solve. Near best performance assures these models are well-trained, while the variance indicates they potentially capture  uniquely different heuristics. \textbf{(2)} \textbf{ZTop} also provides \textbf{\textit{an ensembling technique  to leverage  different models.}}
Conventional ensemble methods (averaging the decisions of different models) do not work well, as we show in the experimental section, since one heuristic typically works only for a group of instances. Averaging all the heuristics may cancel out each heuristic's contribution. So we propose a new ensemble method that applies each model sequentially or in parallel to each test instance and takes the best output per instance. \textbf{ZTop}'s training time 
is identical to the training time of a \textbf{single} model, 
since \textbf{ZTop} selects the models from the same training trajectory. 
During the test time, we have a similar computing overhead as the traditional ensemble. For each test instance, we compute every models' output and take the best one instead of averaging their outputs.  
\textbf{(3)} 
\textbf{ZTop} \textbf{\textit{substantially improves the performance of very different learning frameworks 
on three prototypical CO domains}}, the  hardest unique solution Sudokus \cite{drnet}, the routing problem \cite{attention}, and the graph maximum cut problem \cite{ecodqn}, as well \textbf{\textit{as multi-label classification (MLC),} a machine learning task with a  \textbf{\textit{large combinatorial label space}}},  with \textbf{\textit{zero training overhead.}} These frameworks cover \textbf{\textit{various architecture and learning methods}}: for Sudoku, a modified Long Short Term Memory (LSTM) \cite{LSTM} with weakly supervised learning, for routing problems, a pointer network with attention encoder with reinforcement learning (RL), for  Maximum Cut, structure2vector (s2v) with RL, and for MLC,  Label Message Passing (LaMP) \cite{LaMP} and Higher Order correlation VAE (HotVAE) \cite{hotvae}. 

\section{Related Work}

\textbf{Deep Learning for Combinatorial Optimization (CO):}
Several CO deep learning model architectures leverage the problem structures for learning better heuristics.
Structure2Vector (s2v) \cite{s2v, s2vco} was proposed for graph instances. For each node of the graph, a feature vector can be learned, capturing the properties of itself and its neighbors. A Pointer Network (PN) \cite{pointer} 
computes the permutation of variable length sequence data. This subtle model can naturally capture 
many CO constraints. Graph Neural Networks are also employed  for 
SAT instances \cite{neurosat, gqsat}. A modified LSTM, i.e., each LSTM step's input is multiplied with a constraint graph capturing the Sudoku constraints, performs very well
on weakly-supervised learning to solve challenging Sudoku problems~\cite{drnet}.
Creating  a  large  labeled  dataset  is computationally intractable 
for CO problems. 
So, deep reinforcement learning or weakly supervised learning is used to learn CO problems, ranging from routing problems \cite{pointerrl, attention}, the maximum cut problem \cite{cut1, ecodqn} to the minimum vertex cover problem \cite{cut1, co}. Several approaches 
combine deep models with search algorithms \cite{attention, sample2opt, samplemcts} to further improve the performance.  \textbf{\textit{ZTop can further significantly improve the performance on top of 
these
learning approaches.}}


\textbf{Ensemble Methods in Deep Learning:} Neural network ensembles 
\cite{zhouensemble, ensemble1,ensemble2} have been widely used 
to boost the model's performance in the machine learning community. However, they mainly focus on (weighted) averaging the output of the models and training a set of heterogeneous models 
requiring substantial computational resources. Due to the considerable overhead of traditional ensemble methods, the deep learning community proposes several methods to remedy this issue. ``Implicit'' ensembles \cite{dropout, dropconnect, stochasticdepth, swapout, zoneout, branchout} are proposed as an alternative to traditional ensemble methods 
given their minor overhead in both the training and test phase. One example of ``implicit'' ensembles is Dropout \cite{dropout}, which randomly zeros some hidden neurons during each training step. At the test time, every neuron is kept and scaled by the keeping probability of the training phase. An explanation of dropout is that there are a considerable number of models created by dropping neurons, and these models are implicitly ensembled at the test phase. 
Similar to the dropout mechanism, stochastic depth \cite{stochasticdepth} proposes to randomly drop some layers during the training time to create different models with various depths. These implicit ensemble methods create many shared-weights models and ensemble them in the test phase implicitly. Several works focus on efficiently 
acquiring many good models \cite{snapshot, efficientensemble} to decrease the training cost. Snapshot ensemble \cite{snapshot} leverages the cyclic learning rates \cite{cyclic} to force the model to jump from one local minimum to another local minimum, and
ensemble these models of different local minimum to reduce the training phase overhead. However, cyclic learning rates can potentially damage the model's performance, and it is designed for only convolution neural networks. Another line of research focuses on reducing the test time overhead \cite{modelcompression, distill, meal, newdistill}. Distill \cite{distill} proposes to employ the ensemble of many models as the training label of a single model (similar or smaller size). Learning to improve (L2I) \cite{ensemble_cvrp} proposes a new way to leverage models. L2I employs the model to guide their local search, and they find that taking the best results of many models with small rollout steps is better than one model with much larger rollout steps. However, L2I requires to train these models separately, and the training method is designed specifically for their learning framework.\\ 
Ensemble methods for MLC fall into three groups: the first group is to ensemble binary relevance classifiers~\cite{read2011classifier,wang2016cnn}; the second group is to ensemble label powerset classifiers~\cite{read2008multi, tsoumakas2010random}; and, the third group uses random forest of predictive clustering trees ~\cite{kocev2007ensembles}.
All the ensemble methods above except for L2I rely on (weighted) averaging the outputs of many models. 


\textbf{\emph{Algorithm Portfolios,}} used by the CO community  \cite{gomes2001algorithm, msportfolio, satzilla}, ensemble  different  algorithms and solvers to solve a CO problem. In particular, 
 \textbf{\emph{restart portfolios (restarts)}}  are  widely used in  the SAT community \cite{restart, Biere2018}, since they are an effective way of combating long and heavy tailed runtime distributions~\cite{phase}. They are also used in  stochastic optimization to solve non-convex problems \cite{dick2014many, gagliolo2007learning}. \textbf{\emph{Restarts}} periodically restart an algorithm, de facto trying   different (pseudo) randomized heuristics, with low  heuristic-switching cost.



\textbf{\emph{Restarts} in Deep Learning (DL):}
As an  example of \emph{restarts} in DL, 
 the learning rate is reset based on some manually designed metrics to speed up the training or improve the performance  \cite{lrrestart, cyclic}.  DRNets (Chen et al., 2020) also employ restarts in the optimization phase to improve performance,  with different random seeds. Nevertheless restarts are not often employed in the deep learning community.
 
 
  \textbf{\textit{A key feature that differentiates ZTop is that it  
 leverages different models from the same training trajectory  
  in the test phase to improve performance, with zero training overhead.}}
  

\section{\textbf{ZTop}, a Novel Ensemble Method}
\textbf{ZTop} is a novel ensemble method that 
selects a set of diverse models for learning to solve CO problems  incurring zero training overhead, inspired by the \emph{restart portfolios}. The selected models  are  applied  sequentially or in parallel to solve the test instances, taking the best output per instance. \textbf{ZTop} can be used on top of any weakly supervised or reinforcement CO  learning framework, or even in some cases on top of  a  supervised framework, to significantly improve the performance, with zero training overhead and the same test overhead as traditional ensemble methods. 


Due to the discrete nature of many CO problems, an oracle heuristic, i.e., a heuristic that works for almost every instance, is  unlikely 
to exist. In the CO community, \emph{restarts} are widely employed to remedy this issue by periodically switching to a new heuristic. 
We postulate that it is also difficult for a single deep learning model to perfectly solve all kinds of instances. 
 Surprisingly, we have observed that several models acquired through the \textbf{same} training trajectory contain quite diverse heuristics. These models have similar validation performance, but the subsets of the validation set they perform well vary significantly. 
It is then promising to select several models from the same training trajectory instead of a single one.

\textbf{Selecting diverse models from the same training trajectory}. The goal is to select several models with similar validation performance, but they should perform well on different subsets of the validation set. Often, in CO problems, we naturally have a metric $\phi$ to validate the model's performance, e.g., the average length of the route in the TSP problems.  Surprisingly, selecting top-k models with respect to the metric $\phi$ on the validation set can achieve the goal. The resulting set of models have very comparable performance as the optimal set of models found by enumerating all the possible combinations, as we show in the experimental section. In terms of the training time overhead, we mark the additional training operation of \textbf{ZTop} in \textcolor{blue}{blue} in the Alg.~\ref{alg:mostdiff}. We only need to store each model's validation performance along with its parameters. This additional operation incurs zero training cost since traditional model selection also needs to validate these models.

\begin{algorithm}[t]
   \caption{\textbf{ZTop} method workflow}
   \label{alg:mostdiff}
\begin{algorithmic}
   \STATE {\bfseries Input:} Split Dataset $D_{{train}}$,~$D_{{val}}$,~$D_{{test}}$, performance metric $\phi$, number of ensemble models $n$
   \FOR{epoch$=1$ {\bfseries to} $max\_epoch$}
     \STATE Update model's parameters using $D_{train}$
     \STATE Test model on $D_{val}$, \textcolor{blue}{save the test result along with model's parameters in $val\_res$ object}
   \ENDFOR
     \STATE Compute the performance metric $\phi$ based on $val\_res$. Select the top n models, denoted as $M_1, \ldots, M_n$, w.r.t $\phi$.
    \FOR{$i=1$ {\bfseries to} $n$}
     \STATE Test model $M_i$ on $D_{test}$, and save the result as $res_i$
    \ENDFOR
    \FORALL{$x \in D_{test}$}
    \STATE Take the best output of $x$ in $res_{i, i=1\dots n}$
    \ENDFOR
    
\end{algorithmic}
\end{algorithm}

How to leverage these models is another issue. Each model can be viewed as a heuristic, and different heuristics typically work for different instance groups.  So averaging these models' output may cancel out each heuristic's contribution. 

\textbf{Ensembling models at test time.} \textbf{ZTop} applies its models sequentially or in parallel to solve each test instance and selects the best output per instance (see Alg.~\ref{alg:mostdiff}). 
For  weakly-supervised or reinforcement learning (RL) settings, it is natural to have a metric, 
e.g., the reward used in RL, to select the best output. Consider using a deep model to solve Sudoku. We employ ZTop's models to solve test Sudoku instances sequentially. If any model solves a test instance, no need to try other models. 
Note that each test instance is only fed once into each model, 
so our test time overhead is identical to that of traditional ensemble methods.

\section{Experiments}

We show the performance of the \textbf{ZTop} approach on various CO problems learning frameworks and two MLC learning frameworks. 

\subsection{General Experiment Settings}

\textbf{Baselines.} \textbf{ZTop} has zero training overhead, thus we compare it with methods that share similar training overhead with us. The first baseline is the conventional machine learning model selection scheme (denoted as \textbf{Single}), i.e., it selects a single model with the best validation performance. Note here, the learning framework's training strategy consists of proper implicit ensemble methods. So we are also comparing \textbf{ZTop} with these implicit ensemble methods.  
In terms of leveraging the models, we also compare \textbf{ZTop} with the conventional ensemble method, i.e., averaging the output of all models. We denote this baseline as \textbf{Average}: picking top models w.r.t metric $\phi$ and averaging the outputs of them. The last method is our method \textbf{ZTop}: selecting top models w.r.t metric $\phi$ and taking the best output per instance. 


\textbf{Settings for the learning framework.}
\textbf{ZTopping} a given learning approach  involves adapting released codes and 
following the original paper's settings (for training/test instances, train models, and test models' performance).

\subsection{Hardest unique solution Sudoku}
\begin{figure}[h!]
\centering
\includegraphics[width=0.3\textwidth]{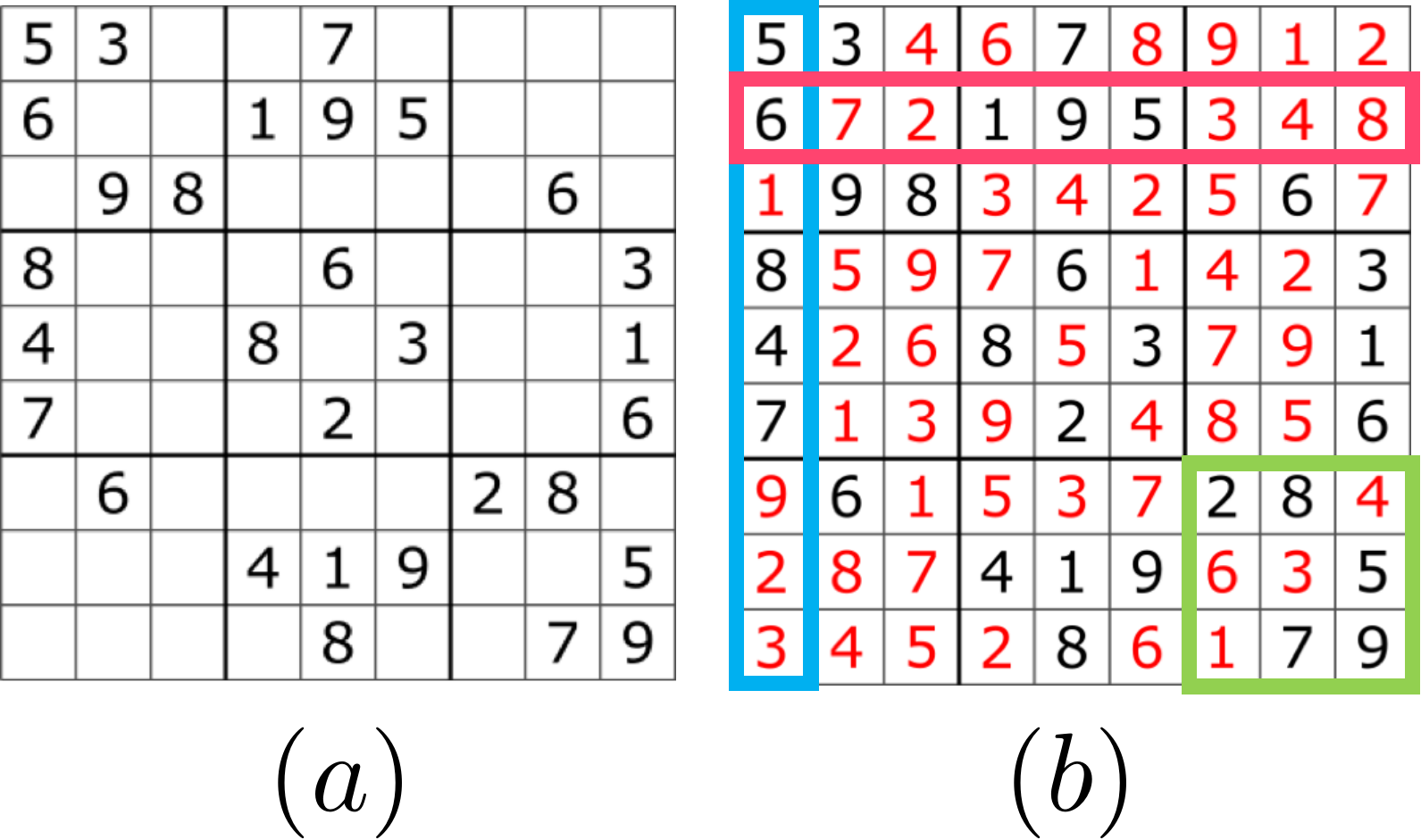}
\caption{\textbf{Sudoku:} (a) a $9\times9$ Sudoku with $30$ hints. (b) The solution to (a); the three rectangles represent the three types of constraints (no repetition of digits per row, column, or block).}
\label{fig:sudoku}
\end{figure}

\begin{table*}[t]
    \centering
    \begin{tabular}{llllllll}
        \toprule
        Single (0) & Single (10) & \# Ensemble Models &  Average (0) & \textbf{ZTop} (0)  & Average (10) & \textbf{ZTop} (10)\\
        \midrule
       0.289	& 0.387& 3&0.364	& \textcolor{blue}{\textbf{0.549}} & 0.569 & \textcolor{blue}{\textbf{0.588}}	\\
            	&   &5& 0.443	& \textcolor{blue}{\textbf{0.652}}  & 0.651	& \textcolor{blue}{\textbf{0.707}}   \\
         & &10	& 0.469	& \textcolor{blue}{\textbf{0.753}} &  0.670	& \textcolor{blue}{\textbf{0.815}}   \\
       &  &20&	 0.475	& \textcolor{blue}{\textbf{0.813}} &  0.452	& \textcolor{blue}{\textbf{0.867}}  \\
        \bottomrule
    \end{tabular}
    \caption{\textbf{ZTopping DRNets, on 1,000 17 hint Sudokus.} DRNets is the state of the art for unsupervised Sudoku~\cite{drnet}. Sudoku accuracy of $17$ hint Sudokus for single and ensemble models, under two test modes: 0 and 10 optimization steps.The number in parentheses next to the modality is the optimization step used in the test mode. We mark ZTop's results in \textcolor{blue}{blue} and \textbf{bold} the best results. ZTop significantly improves the DRNets' performance and outperforms the traditional averaging ensemble method on the two test modes. }
    \label{tb:sudoku0}
\end{table*}


Sudoku is an NP-hard combinatorial number-placement puzzle on an $n \times n$ board. We focus on the Sudoku problems on $9 \times 9$ boards, as they are most widely studied. The object is to complete  a $9\times9$ board, pre-filled with several hints (numbers pre-assigned to cells), with numbers $1$ to $9$. In a  Sudoku, each row, column , and pre-defined $3\times3$ box cannot have a repeated digit (see Fig.~\ref{fig:sudoku}). It has been  showed that $17$ is the minimum number of hints for which  a Sudoku  has a unique solution \cite{17sudoku}. So we focus on solving $9\times9$ Sudokus with 17 hints.

We select DRNets \cite{drnet} as the learning framework since it is  the state-of-the-art weakly-supervised learning  Sudoku framework, supervised \textbf{only} by the Sudoku rules. It employs continuous relaxation to convert the discrete constraints to differentiable loss functions. The training and test of DRNets only require Sudokus without labels.  


DRNets assign each digit a learnable embedding 
and a tensor  represents a Sudoku instance. A modified LSTM computes the missing digits given the Sudoku tensor as input. We create $100, 000$ Sudoku with $18$ to $25$ hints for training and validating the models and $1000$ Sudoku with $17$ hints for testing the model. The other training/validating/testing  settings also strictly follow the original paper. 

To ensure the model candidates are well-trained, we save the top $100$ models from the training phase in terms of the validation performance. Then we select models based on the performance metric $\phi$,  which in this case is the Sudoku accuracy, i.e., the number of Sudoku this model correctly solved. We do not assign partial credits.  We evaluate the ensemble performance of $3, 5, 10$, and $20$ models.

\textbf{Sudoku Accuracy.} The results are summarized in Table~\ref{tb:sudoku0}. We consider two test modes of the DRNets framework: $0$-step optimization mode and $10$-step optimization mode. Since the loss function of DRNets is derived only from the Sudoku rules, it can still  
be optimized 
for the test instances. Here $0$-step mode fixes the models' parameters while $10$-step mode optimizes the loss function $10$ steps in the test phase. In all the cases, 
\textbf{ZTop} 
substantially outperforms the single model and standard averaging ensemble methods. We only require $10$ models to achieve $75.5\%$ Sudoku accuracy, considerably higher than  the single model's $28.9\%$ Sudoku accuracy,  reaching $86.7\%$ with 20 models and 10 optimization steps. 
These results show that  a set of diverse models are more effective than a single model to capture the varied structure of $17$ hint Sudoku instances.
\begin{figure*}[t]
\centering
\includegraphics[width=0.9\textwidth]{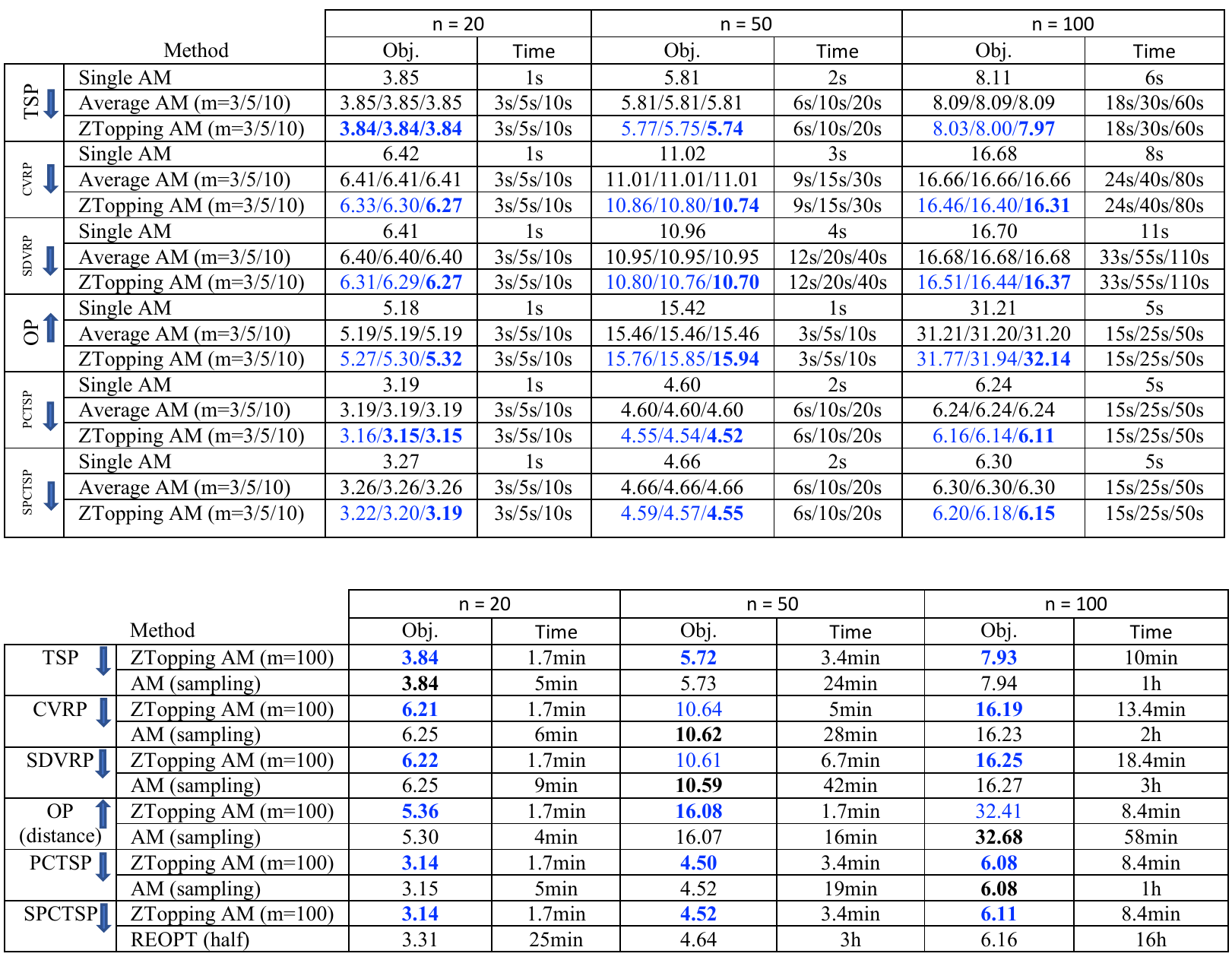}
\caption{\textbf{Performance of \textbf{ZTop} and state-of-the-art AM baselines on $10,000$ instances}.  \textbf{Ztop}'s results are in \textcolor{blue}{blue} and  the best result for each problem is in \textbf{bold}. The variable $n$ represents the number of cities/nodes of the problem. The variable $m$ refers to the number of models used for ensemble. The up arrow next to the problem means  the larger the better the results are while the down arrow means  the smaller the better the results are. The top table  compares all \emph{direct} methods, i.e., no search included. While the bottom table compares \textbf{Ztop} using all the models and the sampling method. Due to the stochastic nature of SPCTSP, AM (sampling) does not work. We compare our method with REOPT (half), a C++ ILS-based algorithm proposed in the original paper. From the top table, we observe that ZTopping AM substantially outperforms AM and the conventional ensemble method using only 3 models for the ensemble.  ZTopping AM using 100 the models  can surprisingly excel the AM (sampling) method with a considerably shorter time in most cases. }
\label{fig:routing}
\end{figure*}
\subsection{A Series of Routing Problems}
Routing problems are intensively studied in the CO community and have many real-world applications \cite{tspapplication}. We consider four routing problems and their variants. \textbf{Traveling Salesperson Problems (TSP):} Given a set of cities and the distance between any two cities, the goal is to find the shortest route that can traverse every city and return to the start city. \textbf{Vehicle Routing Problems (VRP):} This problem \cite{vpr} is a generalized TSP problem. Given  a set of cities and a depot, each city has a demand, i.e., how many people intend to visit the city,  the goal is to find an optimal set of routes to meet all the demands. We focus on two variants of this problem. \textbf{Capacitated VRP (CVRP)}: The total cities' demands  in one route cannot excel a pre-set threshold. \textbf{Split Delivery VRP (SDVRP)}: This variant has the same constraint as CVRP, but here the demand of a city can be split through multiple routes. \textbf{Orienteering Problem (OP) \cite{op}:} We have a set of cities and know the distance between any two cities. Each city contains a prize. The object is to find a route whose length cannot surpass a threshold and we want to maximize the summation of the cities' prizes  in this route. \textbf{Prize Collecting TSP (PCTSP) \cite{pctsp}:} This is a more challenging variant of the TSP problem. Each city is assigned a prize and a penalty. We need to find a route collecting at least a minimum total prize to minimize the route length plus the sum of missed cities' penalty.  \textbf{Stochastic PCTSP (SPCTSP):} This problem is quite similar to the PCTSP problem. The only difference is that the prize of each city is sampled from a fixed distribution. The salesperson only knows the \emph{expected} prize of each city, and the true prize is revealed when visiting a city.  

We select Kool et al.'s \cite{attention} learning framework (denoted as AM) for these routing problems since it is one of the state-of-the-art methods. This framework employs reinforcement learning (REINFORCE algorithm \cite{REINFORCE} with baselines) to learn efficient heuristics unsupervisedly. Thus, the training, validation, and test of this framework only require problem instances.
The model architecture of AM is the pointer network \cite{pointer} with an attention encoder. The graph of the problem is fed into the attention encoder to compute the graph embedding and node embeddings. Then the decoder of the pointer network generates the policy of the next action.

The solution generation process of AM is \emph{incremental}, i.e., the model selects one node (city) per step conditioned on all the previous actions. The data generation and training processes are strictly following the original paper's settings \cite{attention}. We select the best $100$ models from the training phase. We evaluate the ensemble performance of $3, 5$, $10$, and all the $100$ models. The metric we used for each problem is illustrated below.

\textbf{Results.} We introduce one more baseline for the AM method since it proposes a sampling method based on the model's output. This method (denoted as \textbf{AM (sampling)}) can generate better solutions with considerable time. AM (sampling) samples $1280$ solutions and report the best result for each problem. We now introduce the metrics (Obj. column of the table) used to evaluate the performance of each problem. All the metrics we report are averaged across all the test instances. \textbf{TSP}: length of the route, \textbf{CVRP, SDVRP:} length of the route that meets the capacity constraint, \textbf{OP}: the summation of prize collected in the route that meets the length constraint and \textbf{PCTSP, SPCTSP:} length of the route plus the penalty of missing cities. The results are summarized in Fig~\ref{fig:routing}. The top table compares all the \emph{direct} methods, i.e., they do not use any search to generate solutions. We can observe that \textbf{ZTop} significantly outperforms the AM learning framework even with only  $3$ models. We also compare the AM (sampling) method with our \textbf{ZTop} using all (100) models and the results are showed in the bottom table. In most cases, our method can also outperform the AM (sampling) method with a considerably shorter time. In the CVRP ($n$ = 100) problem, we achieve the mean length of route $16.19$ in $13.4$  minutes while AM (sampling) requires $2$ hours to achieve only $16.23$. In the SPCTSP ($n$ = 100) problem, we achieve $6.11$ in $8.4$ minutes while REOPT (half) algorithm requires a substantially long time of $16$ hours to achieves a poorer performance $6.16$. We also observe that averaging ensemble method sometimes even decreases the performance, e.g., the objective drops from $31.21$ to $31.20$ in the OP ($n=100$) problem. Moreover, in most cases, averaging ensemble method only achieves minor improvements.   

\begin{figure*}[t]
\centering
\includegraphics[width=0.95\textwidth]{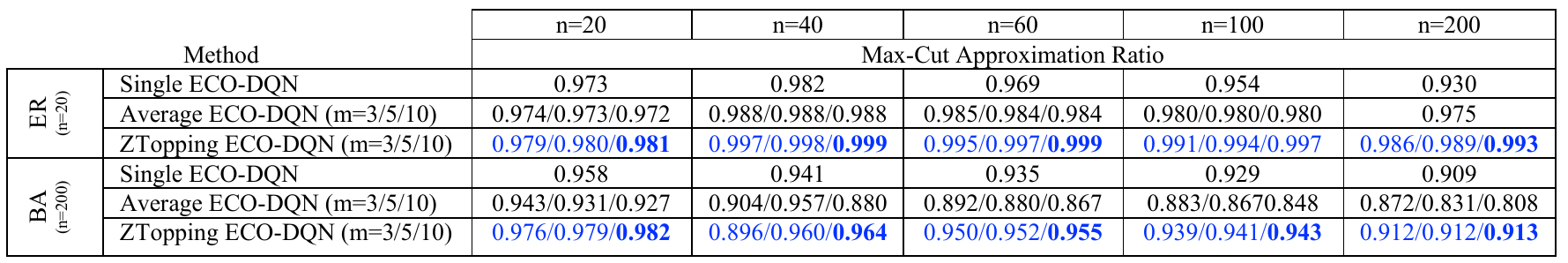}
\caption{\textbf{Approximation ratios of ZTop and ECO-DQN baselines ($100$ instances)}. \textbf{ZTop}'s results are in \textcolor{blue}{blue} and the best results are in  \textbf{bold}. ER and BA refer to two types of the random graphs.  $m$ represents the number of models used for the ensemble. The variable $n$ on the left and top represents the number of vertices of the training set graphs and the test set graphs respectively. ZTopping ECO-DQN substantially improves the ECO-DQN and outperforms the conventional averaging ensemble method in all the cases. }
\label{fig:maxcut}
\end{figure*}
\subsection{Graph Maximum Cut Problem}

The graph maximum cut problem is a classical problem in graph theory. The cut of a graph is a partition of the vertex set into two complementary sets and the number of edges or the sum of edges' weight between these two sets is denoted as the cut's capacity. The maximum cut is the cut that has a maximum capacity.  

We employ the state-of-the-art ECO-DQN \cite{ecodqn} as the learning framework for the maximum cut problem. ECO-DQN leverages reinforcement learning (Deep Q learning \cite{dqn}) to  learn useful heuristics unsupervisedly. So, like the AM learning framework, we do not require any solution to train the model. 
The model architecture of ECO-DQN is Message Passing Neural Networks (MPNN) \cite{MPNN}. It represents each vertex as an embedding. Several rounds of message passing  update the vertices' embeddings. The probability distribution over all the actions is computed through a readout function. 
\begin{figure*}[t]
\centering
\includegraphics[width=0.95\textwidth]{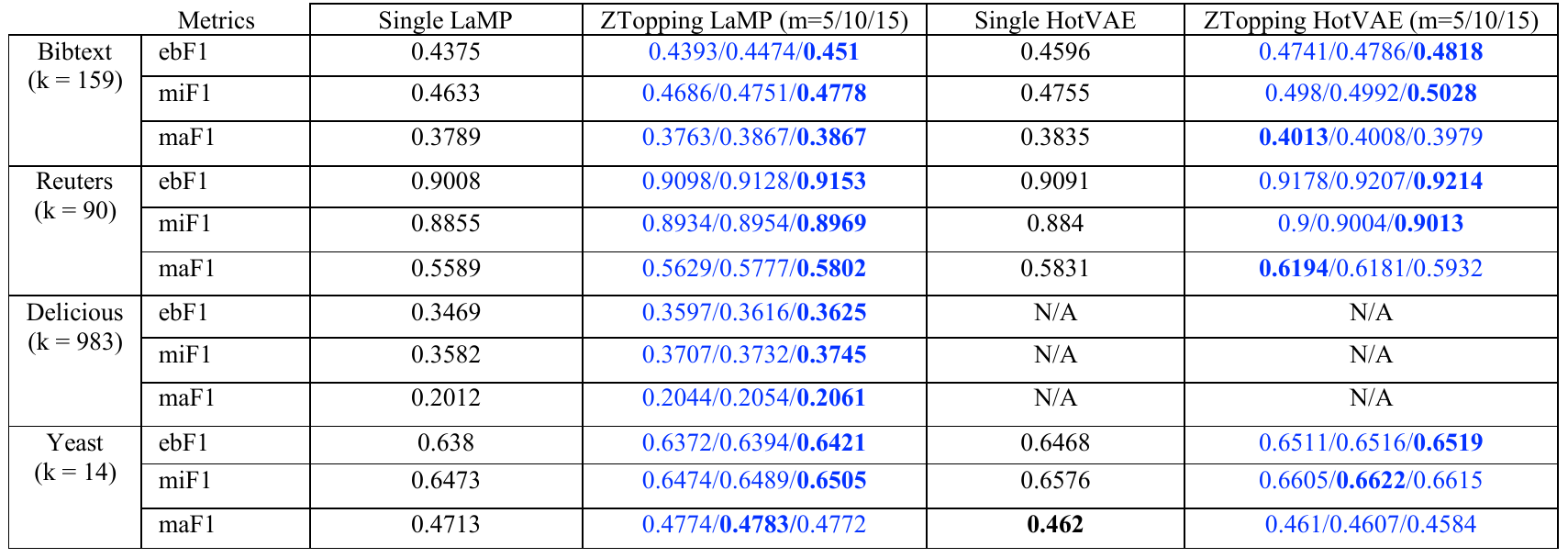}
\caption{\textbf{Multi-label classification.} The performance of \textbf{ZTop} and LaMP/HotVAE on different datasets. \textbf{ZTop}'s results are in \textcolor{blue}{blue} and the best result for each dataset is in \textbf{bold}. $m$ is the number of models used for the ensemble. $k$ is the number of labels. We report ebF1, miF1 and maF1 scores of these approaches. We observe that \textbf{ZTop} significantly improves single LaMP or HotVAE on all the datasets except for Yeast.  \textbf{ZTop} only improves slightly on Yeast since  the dataset is relatively simple, with only 14 labels.  }
\label{fig:mlc}
\end{figure*}
The ECO-DQN proposes the \emph{test-time exploration} concept where their policy is to improve a given solution instead of constructing a solution from scratch. Their original evaluation protocol is to generate $50$ random solutions on the fly, employ the model to improve them and pick the best one. We generate $50$ random solutions in advance and ensure every method is improving the same $50$ solutions. We select $50$ models uniformly from the last quarter of training epochs. We evaluate the ensemble performance of $3, 5,$ and $10$ models. We report the approximation ratio averaged over all the test instances. 

\textbf{Approximation Ratio.} We consider two types of random graphs used in the original paper \cite{ecodqn}. One is Erdős-Rényi (denoted as \textbf{ER}) \cite{er} , and the other is Barabasi-Albert (denoted as \textbf{BA}) \cite{ba} with the edge weights belong to $\{0, 1, -1\}$.  We select two worst-performance cases of the original paper to show ZTopping can improve the ECO-DQN learning framework. 
One case is to train the model on ER ($n=20$) graphs and tested on ER ($n=20,40,60,100,200$) graphs where $n$ refers to the number of vertices, the other is to train the model on BA  graphs and tested on BA ($n=20,40,60,100,200$) graphs.
The results are summarized in Fig~\ref{fig:maxcut}, and ZTop substantially improves the performance of ECO-DQN in all the scenarios. Furthermore, we also observe a similar phenomenon as in the routing problems: the averaging ensemble method decreases the performance. The more models we used for the averaging ensemble, the lower performance it achieves. This phenomenon verifies our assumptions that the averaging ensemble may cancel out each heuristic's contribution.

We show that ZTop can substantially improve the performance of the three learning approaches above. Moreover, the three learning approaches cover two of the most popular architectures: pointer network and s2v, and they also cover both weakly-supervised learning and reinforcement learning algorithms.  The variety of these learning frameworks illustrate the universality of our ZTop method. 
\subsection{Multi-Label Classification} 

Multi-label classification (MLC) 
is the problem of assigning a set of labels to a given object. 
MLC  can be viewed as a combinatorial classification problem, since  the potential MLC's label set is the powerset of the set of single labels. MLC is typically a   
supervised problem,  with  no obvious  unsupervised metric to identify  the best model 
at test time. 
 The MLC models from the same training trajectory are not as heterogeneous as those learned for more standard CO problems, such as TSP or Sudoku. Thus, different MLC models may focus on the object's different label subsets  and therefore an averaging ensemble method is a good  strategy.
So, 
ZTop selects top validation performance models and  averages their outputs, 
in contrast to other CO problems with well defined unsupervised metrics.

We  tested ZTop on the two top state-of-the-art learning approaches for MLC, Label Message Passing (LaMP) \cite{LaMP} and Higher Order correlation VAE (HotVAE) \cite{hotvae} and  on four datasets: Bibtext~\cite{katakis2008multilabel}, Reuters~\cite{lewis2004rcv1}, Delicious~\cite{pmlr-v28-bi13} and Yeast~\cite{nakai1992knowledge}. 
ZTop trains a single MLC architecture and selects $n$ models with the lowest validation losses. ZTop runs inference with the $n$ models and averages their sigmoid scores for the final prediction probabilities. We consider  ZTop ensembles with  5, 10, and 15 models. 
To our knowledge, there are no other deep learning ensemble methods for MLC.

\textbf{F1 scores.} We report three kinds of F1 scores: ebF1, miF1 and maF1 scores. The main results are summarized in Fig.~\ref{fig:mlc}. We can observe that ZTopping LaMP/HotVAE can significantly improve the LaMP/HotVAE on all the datasets except for Yeast. Yeast is a simple dataset with only $14$ potential labels, so \textbf{ZTop} can only improve a little on that dataset. 
\subsection{Ablation Study}
Herein we show how close ZTop (k models) 
is to the optimal k-model ensemble, 
computed by enumerating all possible k-model combinations, out of the  top 100 validation models.
\begin{figure}[h]
\centering
\includegraphics[width=0.44\textwidth]{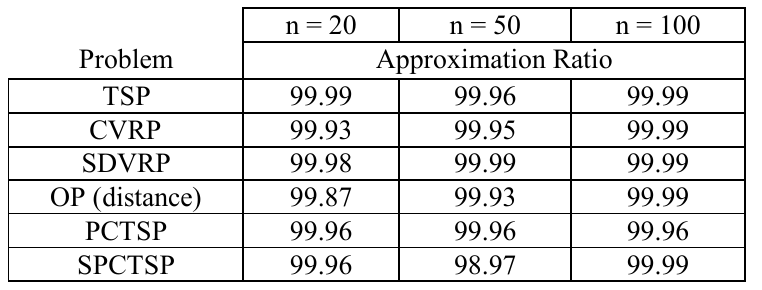}
\caption{The approximation ratio of performance between ZTop (3 best validation models wrt $\phi$) and the optimal 3  models (out of top 100 validation models). $n$ = number of graph vertices. } 
\label{fig:approx}
\end{figure}

We only consider ensembles of size 3 
since enumerating all the combinations 
requires considerable time. We compute the optimal 3-model ensemble (out of the  top 100 validation models), for  
the different routing problems. 
The approximation ratios (Fig~\ref{fig:approx}) are  close to $1$, the worst case is  $98.97$, which shows how  
\textbf{ZTop}'strategy is near-optimal, given the top 100  validation models, implicitly selecting good and  diverse models (heuristics) for these combinatorial problems.


\section{Conclusion}
We introduce the Zero Training Overhead Portfolio (\textbf{ZTop}) method, a simple yet effective model selection and ensemble mechanism to select a set of diverse models for learning to solve combinatorial problems. We demonstrate how ZTopping  substantially improves the current state of the art deep learning frameworks on 
the hardest unique-solution Sudokus, a series of routing problems, the graph maximum cut problem, and  multi-label classification.


\section{Acknowledgements}

This research was supported by NSF awards CCF-1522054
(Expeditions in computing) and CNS-1059284 (Infrastructure). We 
thank Di Chen, Johan Bjorck and Ruihan Wu for their valuable feedback.



\bibliography{example_paper}
\bibliographystyle{icml2021}
\clearpage
\section{Appendix}

\subsection{Ablation Study}
\begin{figure}[h]
\centering
\includegraphics[width=0.45\textwidth]{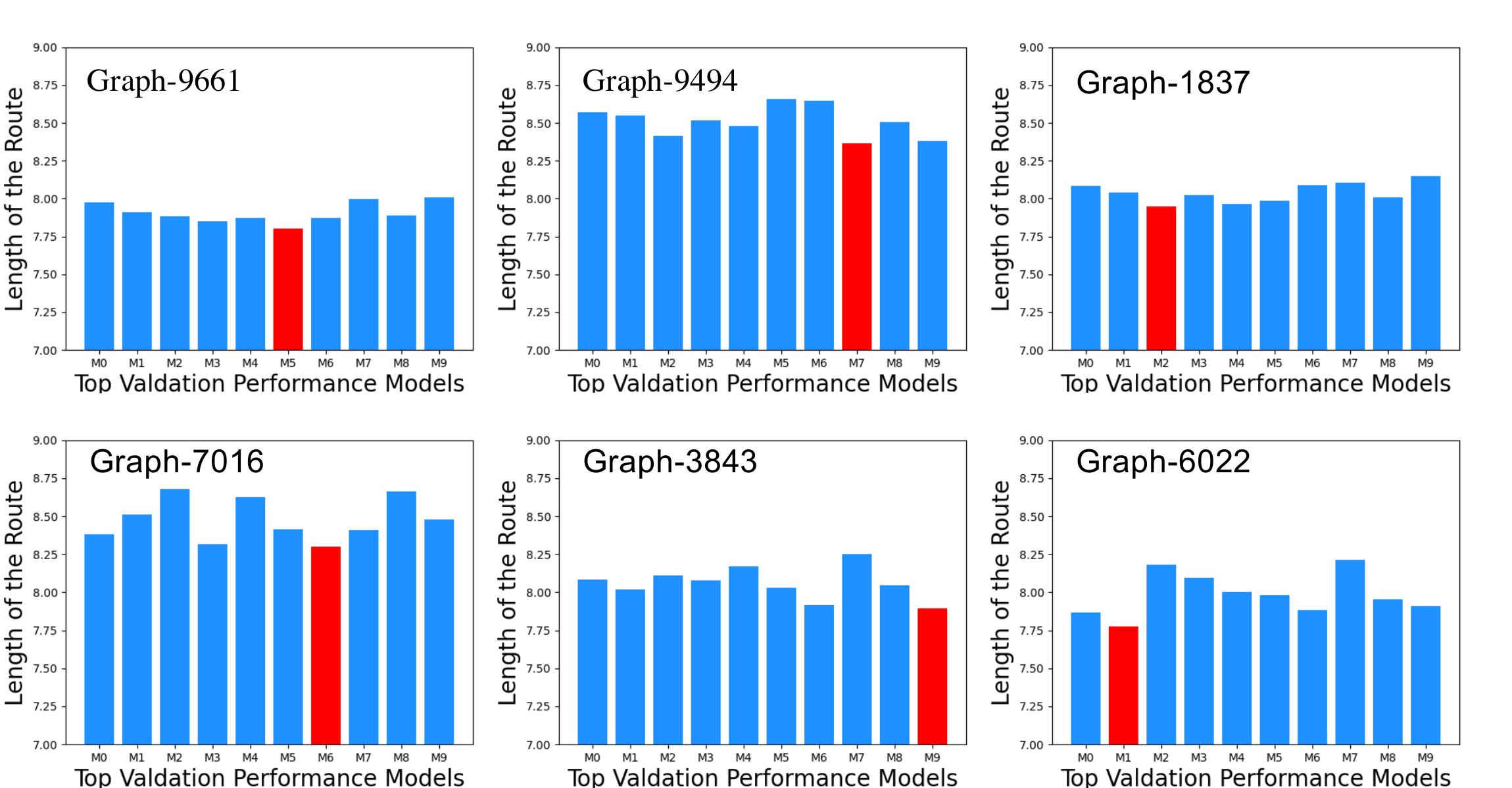}
\caption{We randomly sample $6$ graph instances from the validation set of the TSP-100 problem. We then test the top-10 validation models on these and report their results. One bar corresponds to one model. The red bar means this model performs the best. }
\label{fig:variance}
\end{figure}
We randomly sample $6$ graphs from the $10, 000$ TSP-100 validation instances to show how different these models perform on the validation set. The top-10 validation models are tested on the $6$ graphs, and we report their results in Fig.~\ref{fig:variance}. The models are sorted on x-axis based on the validation performance, and the left first model is the best validation performance model. We can see that the best model of each graph instance is different. For each graph instance, the variance of models' performance is also huge. This verifies our observations.

\subsection{Traditional model selection workflow}

\begin{algorithm}[h]
   \caption{Traditional model selection workflow}
   \label{alg:traditional}
\begin{algorithmic}
   \STATE {\bfseries Input:} Split Dataset $D_{{train}}$,~$D_{{val}}$,~$D_{{test}}$, performance metric $\phi$
   \FOR{epoch$=1$ {\bfseries to} $max\_epoch$}
     \STATE Train model on $D_{train}$.
     \STATE Test model on $D_{val}$ and record the best model $M_{best}$.
   \ENDFOR
   \STATE Test model $M_{best}$ on $D_{test}$.
\end{algorithmic}
\end{algorithm}

We provide the traditional model selection workflow here. This can be compared with our \textbf{ZTop} workflow to show that we incur zero training overhead into the algorithm.

\subsection{More details about the experiments settings}

\subsubsection{Sudoku}

The $17$-hint Sudoku data is public available through here \footnote{https://sites.google.com/site/dobrichev/sudoku-puzzle-collections}. This original site \footnote{http://units.maths.uwa.edu.au/~gordon/sudokumin.php} maintained by the author Gordon Royle is down, so we provide the link above. 



\end{document}